\documentclass{article}

\usepackage{microtype}
\usepackage{graphicx}
\usepackage{booktabs} 

\usepackage[utf8]{inputenc} 
\usepackage[T1]{fontenc}    
\usepackage{hyperref}       
\usepackage{url}            
\usepackage{booktabs}       
\usepackage{amsfonts}       
\usepackage{nicefrac}       
\usepackage{microtype}      
\usepackage{xcolor}         
\usepackage[export]{adjustbox}

\makeatletter
\newcommand*{\defeq}{\mathrel{\rlap{%
                     \raisebox{0.3ex}{$\m@th\cdot$}}%
                     \raisebox{-0.3ex}{$\m@th\cdot$}}%
                     =}
\makeatother

\usepackage{bbm}
\usepackage{apacite}
\usepackage[round]{natbib}

\usepackage{enumitem}
\usepackage{amsfonts}
\usepackage{setspace}
\usepackage{amsmath}
\usepackage{url}
\usepackage{booktabs}
\usepackage{multirow}
\usepackage{wrapfig}
\usepackage{caption}
\usepackage{subcaption}
\usepackage{soul}
\usepackage{dsfont}
\usepackage{rotating}

\usepackage{xspace} 
\usepackage{tikz}
\usetikzlibrary{bayesnet}
\usetikzlibrary{arrows}
\usepackage{color}
\usetikzlibrary{backgrounds}

\usepackage{amsmath,amsfonts,bm}

\def\eqref#1{equation~\ref{#1}}

\def\1{\bm{1}}

\def\rvz{{\mathbf{z}}}

\def\vtheta{{\bm{\theta}}}

\def\vx{{\bm{x}}}

\def\vz{{\bm{z}}}

\DeclareMathAlphabet{\mathsfit}{\encodingdefault}{\sfdefault}{m}{sl}
\SetMathAlphabet{\mathsfit}{bold}{\encodingdefault}{\sfdefault}{bx}{n}

\def\gT{{\mathcal{T}}}

\usepackage{amsthm}
\usepackage{mathtools}

\usepackage{hyperref}

\usepackage[accepted]{icml2024}

\usepackage{amssymb}
\usepackage{float}
\usepackage[capitalize,noabbrev]{cleveref}

\theoremstyle{plain}

\theoremstyle{definition}

\theoremstyle{remark}

\usepackage[textsize=tiny]{todonotes}

\icmltitlerunning{scTree: Discovering Cellular Hierarchies in the Presence of Batch Effects in scRNA-seq Data}

\begin{document}

\twocolumn[
\icmltitle{scTree: Discovering Cellular Hierarchies in the Presence of Batch Effects in scRNA-seq Data}



\icmlsetsymbol{equal}{*}
\icmlsetsymbol{equallast}{†}

\begin{icmlauthorlist}
\icmlauthor{Moritz Vandenhirtz}{equal,yyy}
\icmlauthor{Florian Barkmann}{equal,yyy}
\icmlauthor{Laura Manduchi}{yyy}
\icmlauthor{Julia E. Vogt}{equallast,yyy}
\icmlauthor{Valentina Boeva}{equallast,yyy}
\end{icmlauthorlist}

\icmlaffiliation{yyy}{Department of Computer Science, ETH Zurich, Switzerland}

\icmlcorrespondingauthor{Moritz Vandenhirtz}{moritz.vandenhirtz@inf.ethz.ch}
\icmlcorrespondingauthor{Florian Barkmann}{florian.barkmann@inf.ethz.ch}

\icmlkeywords{Machine Learning, ICML}

\vskip 0.3in
]



\printAffiliationsAndNotice{\icmlEqualContribution} 

\begin{abstract}
We propose a novel method, scTree, for single-cell Tree Variational Autoencoders, extending a hierarchical clustering approach to single-cell RNA sequencing data. scTree corrects for batch effects while simultaneously learning a tree-structured data representation. This VAE-based method allows for a more in-depth understanding of complex cellular landscapes independently of the biasing effects of batches. We show empirically on seven datasets that scTree discovers the underlying clusters of the data and the hierarchical relations between them, as well as outperforms established baseline methods across these datasets. Additionally, we analyze the learned hierarchy to understand its biological relevance, thus underpinning the importance of integrating batch correction directly into the clustering procedure.
\end{abstract}

\section{Introduction}

Recent progress in high-throughput sequencing technologies has enabled single-cell RNA sequencing (scRNA-seq) to emerge as a powerful tool for investigating cellular diversity in various tissues and organisms by providing detailed gene expression profiles across individual cells~\citep{sikkema2022integrated,eraslan2022single}.
Clustering analysis, especially hierarchical clustering, plays an important role in deciphering these data by identifying distinct cellular sub-populations and revealing unknown cell identities and functions~\citep{kiselev2019challenges,osumi2021cell}. However, traditional hierarchical clustering faces challenges with scRNA-seq data, such as amplification biases and high-dimensional spaces, compounded by batch effects from inherent technical variations across experimental batches. 

To address these issues, batch integration methods have been developed that harmonize data across batches, improving the robustness of downstream analyses like clustering and differential expression analysis~\citep{luecken2022benchmarking}. Current best practices for analyzing scRNA-seq data involve a two-step procedure. The first step is dimensionality reduction with batch integration to compress the data, for example with a variational autoencoder~\citep{lopez2018deep, lotfollahi2019conditional}. Then, clustering at different resolutions is performed on the lower dimensional data representation \citep{luecken2019current, hua2019case}. Recently, Tree Variational Autoencoders (TreeVAE)~\citep{Treevae} has demonstrated substantial improvements in clustering performance with a combined, end-to-end optimization of hierarchical clustering and representation learning.

Building upon the work of TreeVAE, we extend this framework to address the challenges presented by scRNA-seq data.
We propose scTree, a method that integrates hierarchical clustering with batch correction techniques to enhance the clustering of scRNA-seq data. Additionally, we introduce a splitting rule that is able to capture the imbalanced clusters in the data. Our approach identifies the inherent hierarchical structure of cellular populations while simultaneously mitigating batch effects, thereby enabling a more precise understanding of cell types and states. By jointly optimizing hierarchical clustering and batch-integrated representation learning within the VAE framework, we offer a powerful tool for dissecting complex cellular landscapes and unraveling the intricacies of biological systems at a single-cell resolution. To the best of our
knowledge, this is the first work that explores hierarchical clustering with VAEs trained jointly with batch integration for scRNA-seq data.

\textbf{Our main contributions} are as follows: i) We propose an extension of TreeVAE to scRNA-seq data that simultaneously corrects for batch effects and learns a binary tree to mimic the hierarchies present in the data. Additionally, we propose a novel splitting rule, removing the bias towards balanced clusters. ii) We evaluate our method on seven different datasets and compare it to three baselines to demonstrate its effectiveness. iii) We qualitatively assess the learned hierarchy and show the correspondence to the underlying biological systems.
\begin{figure*}
    \centering
    \includegraphics[width=1\textwidth]{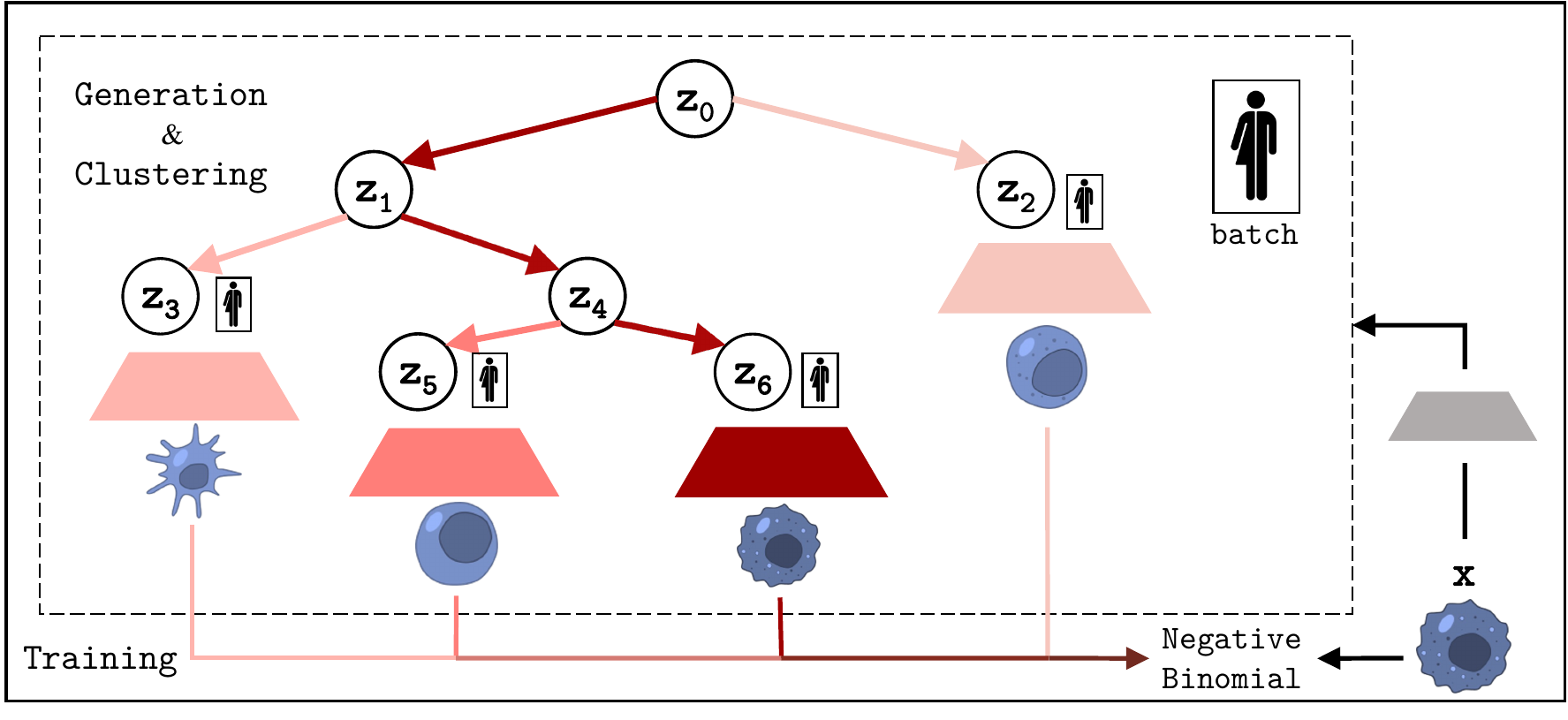}
    \caption{Schematic overview of the proposed method. The input $\vx$ is passed through an encoder to be consequently reconstructed through a tree-shaped process. The process consists of probabilistically going left or right in each node, followed by a nonlinear transformation on the embedding $\vz_i$. The cluster-specific decoders take as input their leaf-embedding and batch information, and reconstruct the gene count parameters of the negative binomial distribution.}
    \label{fig:method}
\end{figure*}

\section{Single-Cell Tree Variational Autoencoders} 
\label{sctree}

We propose scTree, a VAE-based method that uncovers hierarchical structures in single-cell RNA sequencing data. We build upon the recently proposed TreeVAE~\citep{Treevae}, described in \cref{treeVAE}, and extend it to learn a structured latent space corrected for batch effects, thereby enabling the discovery of cell types (and subtypes) in an unsupervised way. scTree improves upon TreeVAE by (a) defining a new reconstruction loss for the new data type, (b) integrating batch information into the architecture, and (c) defining a new splitting criterion. Figure~\ref{fig:method} provides a schematic overview of scTree.  

First, to accommodate for the discrete nature of the data representing read counts, we redefine the loss function. Instead of assuming Bernoulli or Gaussian data, we now assume a Negative Binomial distribution $\vx \mid \vz_{\mathcal{P}_l}, \mathcal{P}_l \sim {\displaystyle \mathrm {NB} \left(\mu_l\left(\vz_l\right)+\mu_b,\, \vtheta\right)}$, where $\mathcal{P}_l$ stands for the path to leaf $l$ and $\mu_b$ is a batch-specific offset. The cluster-specific gene mean is predicted by the leaf-specific decoder $\mu_l$ from the sample-wise latent leaf embeddings $\vz_l$, while the dispersion parameters $\vtheta$ are learned per gene and remain the same across leaves. This parameterization encourages that samples in each leaf are supposed to have unique characteristics that are being captured by the leaf-specific decoders and, as such, supports meaningful clustering.

A frequent issue in scRNA-seq is the handling of batch effects. While previous hierarchical clustering methods perform batch integration either ante-hoc~\citep{li2020deep} or cluster the data at different resolutions~\citep{luecken2019current, hua2019case}, the gradient-based nature of the clustering in TreeVAE allows for an end-to-end integration of batch effects into the learning process. As such, we add a batch-specific offset $\mu_b$ to each cluster-specific gene mean $\mu_l(\vz_l)$. The batch-specific offset is learned during training by passing a one-hot encoding of the batch through a neural network.

An important question for every divisive clustering algorithm is the finding of an adequate criterion that determines which cluster to split further. For TreeVAE, the split is performed by splitting the leaf with the highest number of samples falling into it, which encodes an implicit bias towards balanced clusters. While this works well for balanced imaging benchmarks, in scRNA-seq data, oftentimes, the target cell types are distributed unevenly. For this reason, we introduce a novel splitting rule that is based on the reconstruction loss. For each leaf, we grow a proposal subtree with leaves $l_1, l_2$ for $10$ epochs and compute the average difference in reconstruction loss $\lVert\log p(\vx \mid \vz_{\mathcal{P}_{l_1}}, \mathcal{P}_{l_1}) - \log p(\vx \mid \vz_{\mathcal{P}_{l_2}}, \mathcal{P}_{l_2})\rVert_1$ over the dataset it was trained with. The bigger the difference, the more specialized the leaves have become, indicating that a meaningful split has been found. Thus, the proposal subtree with the highest average difference in reconstruction loss is selected and trained for a longer time.
\begin{table*}[!htb]
\vspace{-0.2cm}
\small
\centering
\caption{Hierarchical clustering performances of scTree compared with baselines. Means and standard deviations are computed across 10 runs with different random model initializations. The best-performing methods are bolded.}
\label{tab:results}
\begin{tabular}{llcccc|cc}
\toprule
 &  & NMI ($\uparrow$) & ARI ($\uparrow$) & DP ($\uparrow$) & LP ($\uparrow$) & NMI$_\text{batch,cluster}$  & NMI$_\text{batch,celltype}$\\
Dataset & Method &  &  &  &  &  & \\
 \hline \rule{0pt}{0.9\normalbaselineskip}
\multirow[c]{5}{*}{Pancreas} & Agg & $0.75$\scriptsize$\pm0.00$ & $0.52$\scriptsize$\pm0.00$ & $0.89$\scriptsize$\pm0.00$ & $0.90$\scriptsize$\pm0.00$ & $0.36$\scriptsize$\pm0.00$ & \\
 & scVI+Agg & $0.76$\scriptsize$\pm0.01$ & $0.57$\scriptsize$\pm0.04$ & $\textbf{0.94}$\scriptsize$\pm0.01$ & $0.94$\scriptsize$\pm0.00$ & $\textbf{0.10}$\scriptsize$\pm0.01$ & \\
 & ldVAE+Agg & $0.77$\scriptsize$\pm0.01$ & $0.55$\scriptsize$\pm0.03$ & $\textbf{0.94}$\scriptsize$\pm0.01$ & $\textbf{0.95}$\scriptsize$\pm0.00$ & $0.15$\scriptsize$\pm0.01$ & 0.08\\
 & scTree$_{\#\text{sample}}$ & $0.75$\scriptsize$\pm0.03$ & $0.56$\scriptsize$\pm0.08$ & $0.90$\scriptsize$\pm0.08$ & $0.91$\scriptsize$\pm0.03$ & $0.21$\scriptsize$\pm0.03$ & \\
 & scTree$_{\text{reconstruction}}$ & $\textbf{0.84}$\scriptsize$\pm0.04$ & $\textbf{0.83}$\scriptsize$\pm0.10$ & $0.92$\scriptsize$\pm0.07$ & $0.94$\scriptsize$\pm0.02$ & $0.13$\scriptsize$\pm0.03$ & \\  \hline \rule{0pt}{0.9\normalbaselineskip}
 \multirow[c]{5}{*}{Lung Atlas} & Agg & $0.69$\scriptsize$\pm0.00$ & $0.46$\scriptsize$\pm0.00$ & $0.54$\scriptsize$\pm0.00$ & $0.71$\scriptsize$\pm0.00$ & $0.47$\scriptsize$\pm0.00$ & \\
 & scVI+Agg & $0.69$\scriptsize$\pm0.01$ & $0.52$\scriptsize$\pm0.03$ & $0.65$\scriptsize$\pm0.02$ & $\textbf{0.75}$\scriptsize$\pm0.01$ & $0.27$\scriptsize$\pm0.01$ & \\
 & ldVAE+Agg & $0.68$\scriptsize$\pm0.01$ & $0.50$\scriptsize$\pm0.04$ & $0.63$\scriptsize$\pm0.03$ & $0.73$\scriptsize$\pm0.01$ & $0.30$\scriptsize$\pm0.01$ & 0.37 \\
 & scTree$_{\#\text{sample}}$ & $0.70$\scriptsize$\pm0.01$ & $0.48$\scriptsize$\pm0.02$ & $0.68$\scriptsize$\pm0.03$ & $0.74$\scriptsize$\pm0.01$ & $\textbf{0.35}$\scriptsize$\pm0.01$ & \\
 & scTree$_{\text{reconstruction}}$ & $\textbf{0.76}$\scriptsize$\pm0.01$ & $\textbf{0.58}$\scriptsize$\pm0.02$ & $\textbf{0.69}$\scriptsize$\pm0.03$ & $\textbf{0.75}$\scriptsize$\pm0.01$ & $0.32$\scriptsize$\pm0.01$ & \\  \hline \rule{0pt}{0.9\normalbaselineskip}
 \multirow[c]{5}{*}{PBMC} & Agg & $0.55$\scriptsize$\pm0.00$ & $0.40$\scriptsize$\pm0.00$ & $0.59$\scriptsize$\pm0.00$ & $0.74$\scriptsize$\pm0.00$ & $0.34$\scriptsize$\pm0.00$ & \\
 & scVI+Agg & $0.67$\scriptsize$\pm0.02$ & $0.56$\scriptsize$\pm0.04$ & $0.72$\scriptsize$\pm0.02$ & $0.82$\scriptsize$\pm0.02$ & $\textbf{0.05}$\scriptsize$\pm0.00$ & \\
 & ldVAE+Agg & $0.69$\scriptsize$\pm0.02$ & $0.57$\scriptsize$\pm0.04$ & $0.72$\scriptsize$\pm0.03$ & $0.82$\scriptsize$\pm0.02$ & $\textbf{0.05}$\scriptsize$\pm0.00$ & 0.05 \\
 & scTree$_{\#\text{sample}}$ & $0.69$\scriptsize$\pm0.03$ & $\textbf{0.61}$\scriptsize$\pm0.06$ & $\textbf{0.75}$\scriptsize$\pm0.04$ & $\textbf{0.83}$\scriptsize$\pm0.02$ & $0.06$\scriptsize$\pm0.01$ & \\
 & scTree$_{\text{reconstruction}}$ & $\textbf{0.71}$\scriptsize$\pm0.02$ & $0.56$\scriptsize$\pm0.08$ & $0.63$\scriptsize$\pm0.08$ & $0.73$\scriptsize$\pm0.08$ & $0.06$\scriptsize$\pm0.01$ & \\  \hline \rule{0pt}{0.9\normalbaselineskip}
 \multirow[c]{5}{*}{Retina} & Agg & $0.79$\scriptsize$\pm0.01$ & $0.56$\scriptsize$\pm0.05$ & $0.95$\scriptsize$\pm0.01$ & $0.89$\scriptsize$\pm0.00$ & $0.03$\scriptsize$\pm0.00$ & \\
 & scVI+Agg & $0.83$\scriptsize$\pm0.01$ & $0.55$\scriptsize$\pm0.02$ & $0.96$\scriptsize$\pm0.01$ & $0.94$\scriptsize$\pm0.01$ & $\textbf{0.02}$\scriptsize$\pm0.00$ & \\
 & ldVAE+Agg & $\textbf{0.88}$\scriptsize$\pm0.01$ & $0.74$\scriptsize$\pm0.08$ & $\textbf{0.97}$\scriptsize$\pm0.01$ & $\textbf{0.95}$\scriptsize$\pm0.00$ & $\textbf{0.02}$\scriptsize$\pm0.00$ & 0.02 \\
 & scTree$_{\#\text{sample}}$ & $0.87$\scriptsize$\pm0.03$ & $0.86$\scriptsize$\pm0.10$ & $\textbf{0.97}$\scriptsize$\pm0.01$ & $0.91$\scriptsize$\pm0.02$ & $0.03$\scriptsize$\pm0.00$ & \\
 & scTree$_{\text{reconstruction}}$ & $0.86$\scriptsize$\pm0.13$ & $\textbf{0.87}$\scriptsize$\pm0.18$ & $0.96$\scriptsize$\pm0.04$ & $0.88$\scriptsize$\pm0.14$ & $0.03$\scriptsize$\pm0.00$ & \\ \hline \rule{0pt}{0.9\normalbaselineskip}
\multirow[c]{5}{*}{IHC} & Agg & $0.66$\scriptsize$\pm0.00$ & $0.44$\scriptsize$\pm0.00$ & $0.68$\scriptsize$\pm0.00$ & $0.74$\scriptsize$\pm0.00$ & $0.45$\scriptsize$\pm0.00$ & \\
 & scVI+Agg & $\textbf{0.75}$\scriptsize$\pm0.02$ & $\textbf{0.59}$\scriptsize$\pm0.05$ & $\textbf{0.81}$\scriptsize$\pm0.01$ & $\textbf{0.85}$\scriptsize$\pm0.02$ & $0.14$\scriptsize$\pm0.00$ & \\
 & ldVAE+Agg & $\textbf{0.75}$\scriptsize$\pm0.01$ & $0.54$\scriptsize$\pm0.03$ & $0.78$\scriptsize$\pm0.02$ & $0.83$\scriptsize$\pm0.01$ & $0.14$\scriptsize$\pm0.00$ & 0.16 \\
 & scTree$_{\#\text{sample}}$ & $0.73$\scriptsize$\pm0.02$ & $0.53$\scriptsize$\pm0.04$ & $\textbf{0.81}$\scriptsize$\pm0.01$ & $0.83$\scriptsize$\pm0.02$ & $0.18$\scriptsize$\pm0.02$ & \\
 & scTree$_{\text{reconstruction}}$ & $0.68$\scriptsize$\pm0.07$ & $0.48$\scriptsize$\pm0.12$ & $0.60$\scriptsize$\pm0.08$ & $0.69$\scriptsize$\pm0.07$ & $\textbf{0.15}$\scriptsize$\pm0.01$ & \\ \hline \rule{0pt}{0.9\normalbaselineskip}
\multirow[c]{5}{*}{SCC} & Agg & $0.36$\scriptsize$\pm0.00$ & $0.47$\scriptsize$\pm0.00$ & $0.70$\scriptsize$\pm0.00$ & $0.74$\scriptsize$\pm0.00$ & $0.25$\scriptsize$\pm0.00$ & \\
 & scVI+Agg & $0.34$\scriptsize$\pm0.07$ & $0.42$\scriptsize$\pm0.10$ & $0.66$\scriptsize$\pm0.08$ & $0.71$\scriptsize$\pm0.06$ & $0.07$\scriptsize$\pm0.02$ & \\
 & ldVAE+Agg & $0.46$\scriptsize$\pm0.06$ & $0.56$\scriptsize$\pm0.10$ & $0.75$\scriptsize$\pm0.05$ & $0.77$\scriptsize$\pm0.04$ & $\textbf{0.08}$\scriptsize$\pm0.03$ & 0.08 \\
 & scTree$_{\#\text{sample}}$ & $0.50$\scriptsize$\pm0.09$ & $0.51$\scriptsize$\pm0.10$ & $0.80$\scriptsize$\pm0.06$ & $0.80$\scriptsize$\pm0.05$ & $0.14$\scriptsize$\pm0.05$ & \\ 
 & scTree$_{\text{reconstruction}}$ & $\textbf{0.56}$\scriptsize$\pm0.08$ & $\textbf{0.63}$\scriptsize$\pm0.11$ & $\textbf{\textbf{0.81}}$\scriptsize$\pm0.05$ & $\textbf{\textbf{0.81}}$\scriptsize$\pm0.04$ & $0.11$\scriptsize$\pm0.05$ & \\  \hline \rule{0pt}{0.9\normalbaselineskip}
\multirow[c]{5}{*}{GBM} & Agg & $0.42$\scriptsize$\pm0.00$ & $0.42$\scriptsize$\pm0.00$ & $0.58$\scriptsize$\pm0.00$ & $0.66$\scriptsize$\pm0.00$ & $0.37$\scriptsize$\pm0.00$ & \\
 & scVI+Agg & $0.28$\scriptsize$\pm0.03$ & $0.23$\scriptsize$\pm0.03$ & $0.43$\scriptsize$\pm0.01$ & $0.56$\scriptsize$\pm0.03$ & $0.07$\scriptsize$\pm0.00$ & \\
 & ldVAE+Agg & $0.48$\scriptsize$\pm0.03$ & $0.44$\scriptsize$\pm0.07$ & $0.52$\scriptsize$\pm0.05$ & $0.66$\scriptsize$\pm0.04$ & $0.15$\scriptsize$\pm0.01$ & 0.23 \\
 & scTree$_{\#\text{sample}}$ & $\textbf{0.53}$\scriptsize$\pm0.04$ & $\textbf{0.51}$\scriptsize$\pm0.07$ & $\textbf{0.66}$\scriptsize$\pm0.06$ & $\textbf{0.77}$\scriptsize$\pm0.06$ & $\textbf{0.23}$\scriptsize$\pm0.02$ & \\
 & scTree$_{\text{reconstruction}}$ & $0.52$\scriptsize$\pm0.04$ & $0.47$\scriptsize$\pm0.05$ & $0.60$\scriptsize$\pm0.03$ & $0.69$\scriptsize$\pm0.05$ & $0.22$\scriptsize$\pm0.02$ & \\
\midrule
\bottomrule
\end{tabular}
\end{table*}
\begin{figure*}[!htb]
\centering
\includegraphics[width=\textwidth]{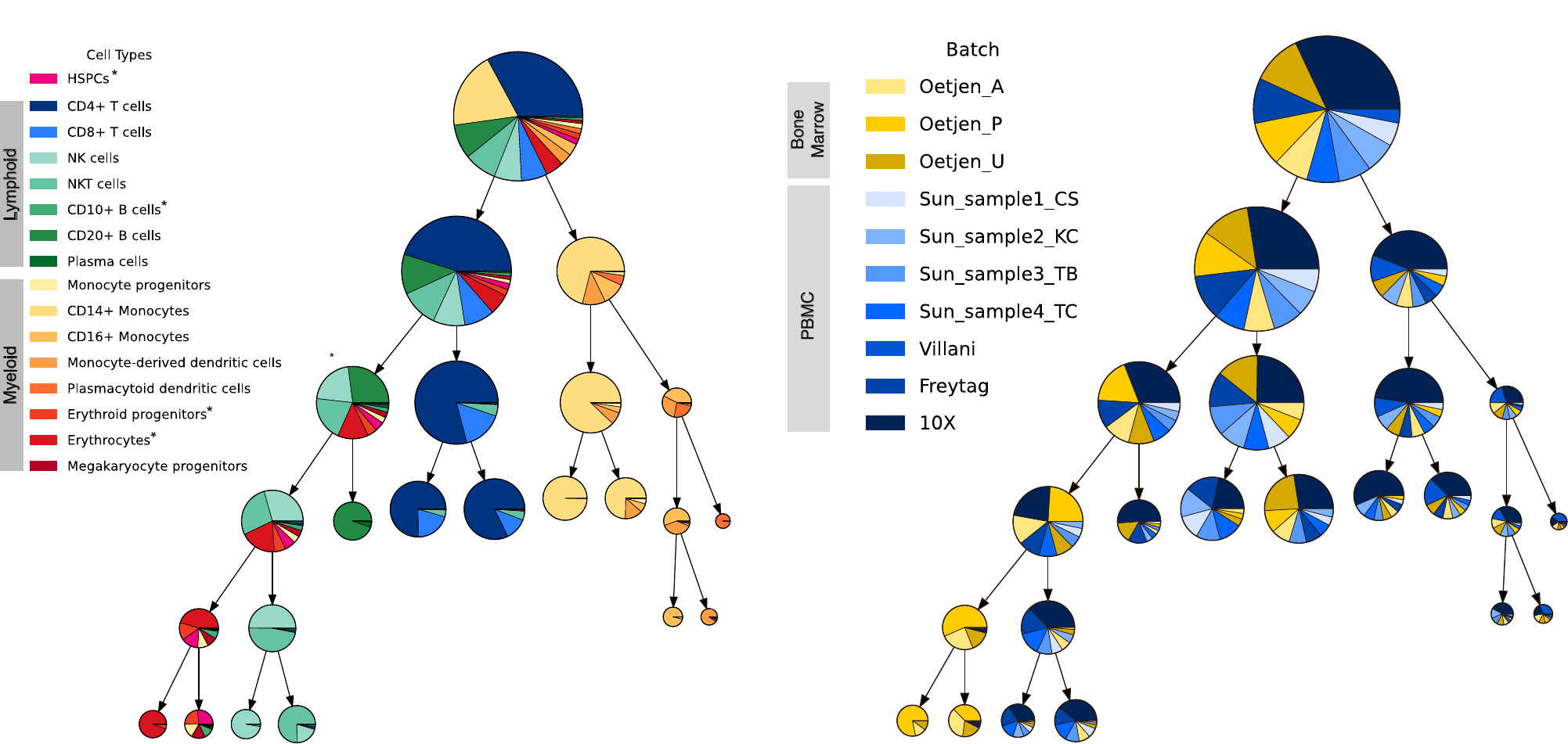}
    \caption{Visualization of hierarchy discovered by scTree on IHC. The size of each node represents the number of cells assigned to it. We exclud empty leaves from the tree. Left: Hierarchy of cell types. Lymphoids, Myeloids and HSPCs have distinct colors. The ``*'' indicates cell types exclusive to the bone marrow samples. Right: Hierarchy of batches. Bone marrow and PBMC batches have distinct colors.
    }
    \label{fig:tree}
\end{figure*}
\section{Experiments}
\label{experiments}
\textbf{Datasets and Metrics} We assess the clustering and batch integration capabilities of scTree using seven distinct scRNA-seq datasets described in \cref{sec:datasets}. Notably, in the IHC dataset~\citep{luecken2022benchmarking}, five cell types are only present in three out of ten batches. Here, we evaluate the methods' capabilities to merge cell types found in all batches without excessively merging cell types that are only present in a few batches.
To evaluate the biological meaningfulness of the hierarchical clustering, we compute the Normalized Mutual Information (NMI) and Adjusted Rand Index (ARI), as well as Dendrogram Purity (DP) and Leaf Purity (LP), as defined by \citep{KobrenMKM17}, using the cell type labels as ground truth. 
We also calculate the NMI of the batch labels and the clustering (NMI$_\text{batch,cluster}$) to assess the batch integration performance. 
We report all metrics for the number of clusters set equal to the number of cell types. 

\textbf{Baselines} We compare scTree to three baselines. Firstly, we employ Principal Component Analysis (PCA)~\citep{pearson1901liii} coupled with Ward’s Agglomerative clustering (Agg)~\citep{Ward1963HierarchicalGT} applied to the first 50 Principal Components (PCs). Additionally, we utilize Agg on the latent representations learned by two commonly used batch integration method for scRNA-seq data: scVI by \citet{lopez2018deep} + Agg and LDVAE by \citet{svensson2020interpretable} + Agg. Additional information pertaining to the baselines and other related work is given in \cref{app:rel}.

\textbf{Implementation Details} 
The model architecture of scTree was determined via datasets that are excluded from this work to prevent biased results. We set the latent dimensions for all VAE-based methods to $10$ and report the results for scTree using both the proposed reconstruction-loss-based and the previous sample-count-based splitting rules. A full list of scTree's architecture can be found in Appendix~\ref{sec:hyperparameters}.

\section{Results \& Discussion}
\textbf{Hierarchical Clustering Results} As evidenced by Table~\ref{tab:results}, on most datasets scTree performs hierarchical clustering on par or better than the baseline methods. Especially for the tumor datasets SCC and GBM, scTree shows promising performance, as evidenced by NMI, as well as DP, where DP also takes the learned hierarchy into account. As the tumor datasets are the ones that pose the most significant challenge regarding batch integration, we interpret these results that scTree performs best when there is a strong need to correct for these patient-specific effects. 
To measure batch integration performance, we analyze the NMI$_\text{batch,cluster}$. While scTree usually has a higher value than the baselines, the interpretation of the values is difficult, as the ground-truth clustering is not independent of the batch labels as shown by NMI$_\text{batch,celltype}$. Thus, we ascertain that the best method is the one closest to this value.
With this in mind, the results suggest that scVI and LDVAE might over-integrate in some cases. Contrarily, scTree's end-to-end batch integration can separate irrelevant batch effects from the ones correlated with clusters, as scTree's clustering performance retains an adequate amount of batch information necessary for a meaningful clustering.

\textbf{Discovery of Hierarchies}
Figure \ref{fig:tree} (left) shows that early in the hierarchy scTree correctly separates Lymphoid and Myeloid cell types present in both PBMCs and bone marrow datasets. At first, it inaccurately allocates many Myeloid cell types exclusive to bone marrow datasets to the Lymphoid branch but then rectifies this at a lower hierarchy level. Notably, scTree segregates bone marrow exclusive cell types without integrating them with other cell types, as seen in the leftmost subtree of Figure \ref{fig:tree} (right).
scTree achieves pure leaves for most cell types and only struggles with accurate separation of CD4+ and CD8+ T cells, as well as the small bone-marrow-specific cell types. This is also evident by the root node embedding as shown in Figure \ref{fig:umap}. This suggests that the encoder is well-suited to accurately split cell types into distinct clusters while not over-integrating batch effects.

\section{Conclusion}
In this paper, we introduced scTree, a new hierarchical clustering method that directly integrates batch correction techniques into the training to enhance the clustering of scRNA-seq data. We have proposed a novel splitting rule based on the reconstruction loss that detects small clusters. We have shown that scTree performs as well as or better than state-of-the-art methods, especially on data with strong batch effects, such as cancer datasets. We presented qualitatively that scTree learns a biologically plausible hierarchical structure, thereby facilitating the exploration and analysis of scRNA-seq data. To the best of our knowledge, scTree is the first work to explore hierarchical clustering with VAEs jointly with batch integration, where we have highlighted the significant potential of such an approach for exciting advancements in the field. Limitations and future work are discussed in Appendix~\ref{app:lim_fut}.

\section*{Code Availablility} An implementation of scTree is available at \url{https://github.com/mvandenhi/sctree-public}. To reproduce all results, we provide \url{https://github.com/mvandenhi/sctree-supplementary-public}.

\section*{Acknowledgements}

MV is supported by the Swiss State Secretariat for Education, Research and Innovation (SERI) under contract number MB22.00047. LM is supported by the SDSC PhD Fellowship \#1-001568-037. FB is supported by the Swiss National Science Foundation (SNSF) (grant number 205321\_207931).

\bibliography{bibliography}
\bibliographystyle{icml2024}

\newpage
\appendix
\onecolumn
\section{Background}
\label{treeVAE}
This section provides an overview of the Tree Variational Autoencoder~\citep[TreeVAE][]{Treevae}. TreeVAE is a hierarchical VAE composed of a tree structure of latent variables, whose structure is learned during training. It thus learns (i) a hierarchical generative model that permits the generation of new samples and (ii) a hierarchical clustering of data points, thus uncovering meaningful patterns in the data, and a hierarchical categorization of samples.

TreeVAE defines a probabilistic binary tree $\gT$, where each node $i$ is characterized by a sample-specific embedding $\rvz_i$. The generative path of a sample is as follows: First, the root node's latent embedding $\rvz_0$ is sampled from a standard Gaussian. From this embedding, the probabilities of going to the left or right child in the tree are computed by a multilayer perceptron. The latent embedding of the selected child $\vz_i$ is sampled from a Gaussian distribution $p_{\theta}(\vz_i \mid \vz_0) = \mathcal{N}\left(\mathbf{z}_{i} \mid \mu_{p, i}\left(\vz_0\right), \sigma_{p, i}^{2}\left(\vz_0\right)\right)$ conditioned on its parent. This routing--transformation process is repeated until a leaf node is reached. Each leaf corresponds to one cluster and includes a decoder through which the observed sample is generated, conditioned on the sample-specific leaf embedding. 
To recover the assumed generative model, the inference model of TreeVAE matches the tree structure. To avoid a posterior collapse of the root, they utilize the trick of LadderVAE~\citep{Snderby2016LadderVA} to learn a bottom-up chain from the sample $\vx$ to the root $\vz_0$ with which the generative model can be guided. 

To optimize the parameters of the generative and inference model and to learn the tree structure, TreeVAE iterates two training steps sequentially: model refinement and tree growing.
During the model refinement, it assumes a fixed tree (starting from a root and two children) and optimizes the Evidence Lower Bound (ELBO):
$\mathcal{L}(\vx \mid \gT):= \mathbb{E}_{q(\vz_{\mathcal{P}_l}, \mathcal{P}_l \mid \vx)}[\log p(\vx \mid \vz_{\mathcal{P}_l}, \mathcal{P}_l)]-\operatorname{KL}\left(q\left(\vz_{\mathcal{P}_l}, \mathcal{P}_l \mid \vx\right) \middle\| p\left(\vz_{\mathcal{P}_l}, \mathcal{P}_l\right)\right), \label{eq:elbo}$
where $\mathcal{P}_l$ denotes the path in the tree from the root to leaf $l$ which has been followed. A more detailed analysis of the ELBO is omitted, but intuitively, the loss consists of two parts: The first term represents the reconstruction loss, which is characterized by a weighted sum over the reconstruction loss of each leaf, where each weight is the probability that the sample reaches this leaf. For each sample encourages that the leaf with the highest probability has the lowest reconstruction loss, which, combined with the cluster-specific decoders, guides the learning of the clusters. The second part is the Kullback–Leibler divergence (KL), which regularizes the learned embeddings, as well as the routing probabilities. 

In the growing step, the leaf with the highest number of assigned samples is split by attaching two new leaves. The new tree is then updated via the model's refinement step. This scheme is repeated until the tree is fully grown. This imposes an inductive bias towards balanced clusters, which is unsuitable for scRNA-seq data where important cell types might be underrepresented.
For a more detailed description of TreeVAE, we refer to their work. 

\section{Related Work}
\label{app:rel}
Hierarchical clustering algorithms are a frequently used technique for unraveling the intricate hierarchical structures inherent in biological data. Agglomerative hierarchical clustering algorithms~\citep{sneath1957application, Ward1963HierarchicalGT,murtagh2012algorithms} treat each data point as a separate cluster and progressively merge these clusters based on their proximity, as defined by a specific distance metric. Diverging from traditional agglomerative techniques, Bayesian Hierarchical Clustering~\citep{heller2005bayesian} introduces a probabilistic framework that utilizes hypothesis testing for cluster merging decisions. Divisive hierarchical clustering algorithms~\citep{kaufman2009finding}, the category which TreeVAE falls into, offer an alternative strategy, starting with a single cluster that encompasses all data points and iteratively dividing it into smaller clusters. The Bisecting-K-means algorithm~\citep{Steinbach2000ACO, Nistr2006ScalableRW} repeatedly applies k-means clustering to divide data into two parts. Relatedly, \citet{williams1999mcmc} learn a hierarchical probabilistic Gaussian mixture model. Further hierarchical probabilistic clustering methods include VAE-nCRP \citep{Goyal2017NonparametricVA, Shin2019HierarchicallyCR} and the TMC-VAE \citep{Vikram2018TheLP}. that use Bayesian nonparametric hierarchical clustering based on the nested Chinese restaurant process (nCRP) prior \citep{hierarchicalpriorBlei} or the time-marginalized coalescent (TMC). 

Various hierarchical clustering algorithms have emerged to address the unique challenges encountered in single-cell RNA sequencing (scRNA-seq) data analysis. \cite{lin2017cidr} proposed Clustering through Imputation and Dimensionality Reduction (CIDR), leveraging imputation techniques within a hierarchical framework to mitigate the impact of dropouts inherent in scRNA-seq data. \cite{morelli2021nested} presented Nested Stochastic Block Models (NSBM) and \cite{zou2021hgc} proposed Hierarchical Graph-based clustering (HGC), both offering methods for hierarchical clustering directly on the k-nearest neighbor graph of cells, bypassing the count matrix. Additionally, scDEF, a method introduced by \cite{Ferreira2022.10.15.512383}, employs a two-level Bayesian matrix factorization model to jointly generate hierarchical clustering and infer gene signatures for each cluster. Notably, among these methods, only scDEF has the capability to handle batch effects. However, it generates a two-level hierarchy rather than a binary tree, posing challenges for comparisons with methods such as scTree.

\section{Model Architecture}
\label{sec:hyperparameters}
As TreeVAE, and therefore also scTree, has a large number of important hyperparameters that need to be tuned, we use a suitable configuration, which is adjusted for the simpler nature of the input data such that it clusters scRNA-seq data in a meaningful way. All experiments were performed on datasets that are not presented in Section~\ref{experiments} to not bias the results. The hyperparameters and their determined values are presented in Table~\ref{tab:hyperparams}.
\begin{table}[htbp]
    \centering
    \begin{tabular}{l|l}
         Latent dimensions& 10  \\
         Bottom-up latent dimensions& 128\\
         Encoder & Linear layer + Batchnorm + LeakyReLU\\
         Decoder & Linear layer \\
         Transformations & 1 Hidden Layer\\
         Routers & 1 Hidden Layer\\
         Kl-annealing & Linear from 0.001 to 1\\
         Subtree training epochs & 100\\
         Intermediate finetuning epochs & 50\\
         Final finetuning epochs & 50\\
         Batch size& 128\\
         Optimizer & Adam\\
         Learning rate & 0.001\\
         Weight decay & 0.00001
         
    \end{tabular}
    \caption{Hyperparameter configuration of scTree}
    \label{tab:hyperparams}
\end{table}

\section{Datasets} \label{sec:datasets}
The first dataset from \cite{ding2019systematic} consists of peripheral blood mononuclear cells (PBMC) sourced from two healthy donors, which were sequenced on seven different sequencing technologies. In this dataset, the primary challenge lies in harmonizing batches originating from different donors and sequencing technologies. The second dataset is the mouse retinal bipolar neuron dataset (Retina) from \cite{shekhar2016comprehensive} which was also used in \cite{lopez2018deep}. Further three dataset, immune human cell dataset (IHC), Pancreas and Lung Atlas, are taken from \cite{luecken2022benchmarking}. In the IHC dataset, five cell types are only present in three out of ten batches. On this dataset, we evaluate the method's capability to merge cell types consistently found in all batches without excessively merging cell types that are only present in a few batches.  The remaining two datasets consist of malignant cells from cancer patients. Specifically, we utilize a glioblastoma dataset (GBM) from \cite{neftel2019integrative} and a squamous cell carcinoma dataset (SCC) from \cite{ji2020multimodal}. In datasets derived from malignant cells, strong patient-specific effects due to genetic differences between cells pose significant challenges for data integration.
All datasets used in this paper are publicly available. For all datasets and methods we used the 4000 most highly variable genes computed with scanpy's $\mathrm{highly\_variable\_genes}$ \citep{wolf2018scanpy} function with default parameters. We used the same preprocessing as proposed by the authors for each dataset and removed all cells not assigned to a cell type. Below, we present a detailed description of the variables of the aforementioned datasets.

\textbf{IHC:} The IHC dataset contains 33,506 cells from 10 different batch from five different studies with 16 unique cell types. The whole dataset is available under \url{https://doi.org/10.6084/m9.figshare.12420968.v8}.

\textbf{PBMC:} The PBMC datasets consists of 30,449 cells from two healthy donors sequenced with six different sequencing technologies (10x Chromium (v2), 10x Chromium (v3), CEL-Seq2, Drop-seq, Seq-Well, and inDrops). Since LDVAE cannot take additional categorical covariates as input, we generated a new batch column from the donor ID and the sequencing technology and used it for all methods. The dataset is available at \url{https://singlecell.broadinstitute.org/single_cell/study/SCP424/single-cell-comparison-pbmc-data}.

\textbf{Pancreas:} The Pancreas dataset consists of 16,382 cells, 9 batches and 14 cell types. The dataset is available at \url{https://figshare.com/ndownloader/files/24539828}.

\textbf{Lung Atlas:} The Lung Atlas dataset consists of 32,426 cells, 16 batches and 16 cell types. The dataset is available at \url{https://figshare.com/ndownloader/files/24539942}.

\textbf{Retina:} The Retina dataset consists of 19,829 cells, 2 batches and 15 cell types. The dataset is available at \url{https://github.com/broadinstitute/BipolarCell2016}.

\textbf{GBM:} The GBM dataset contains 6,855 malignant cells from 27 different patient. The authors annotated four cellular stats (AC-like, MES-like, NPC-like and OPC-like) describing intra-patient heterogeneity.  We removed all cycling cells from the dataset to ensure that we only retain cells assigned to one of the stats. All samples were sequenced using Smart-seq 2. The dataset is available at \url{https://singlecell.broadinstitute.org/single_cell/study/SCP393/single-cell-rna-seq-of-adult-and-pediatric-glioblastoma}.

\textbf{SCC:} The SCC dataset encompasses 10,529 malignant cells from 8 patients. The authors assigned cells to three cellular stats (Basel, Differentiated and TSK). We again removed all cycling cells. The dataset is available at \url{https://www.ncbi.nlm.nih.gov/geo/query/acc.cgi?acc=GSE144240}.

\section{Visualization of the datasets}
\label{app:embedding}
\begin{figure*}[!h]
\centering
\includegraphics[width=\textwidth]{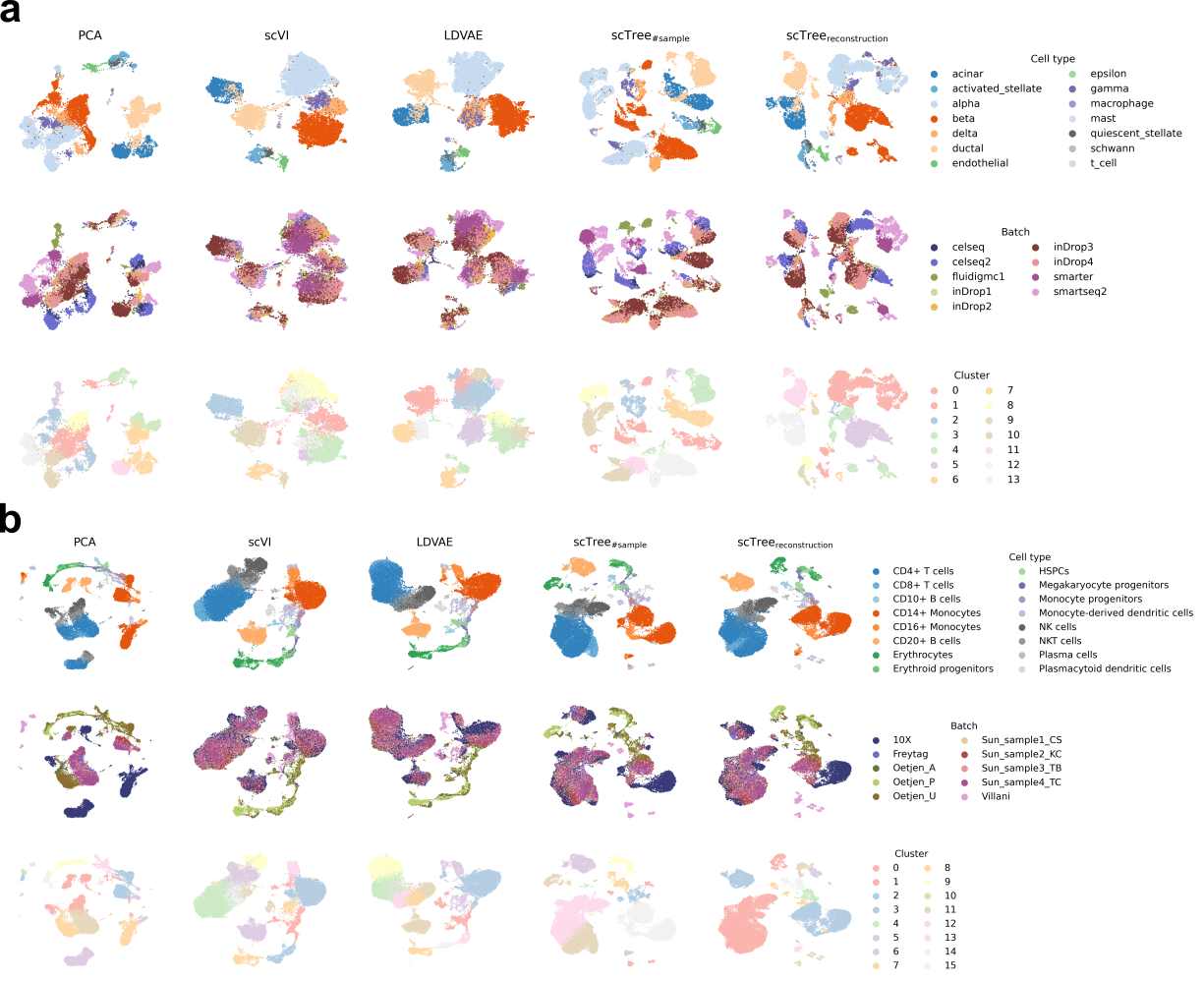}
    \caption{The plots show  uniform manifold approximation and projections based on the first 50 PCs computed on the log-transformed normalized gene expression, the latent representations of scVI and LDVAE, and the Root node representation of scTree with both splitting rules of the Pancreas (a) and the IHC (b) datasets. The plots are colored by cell type (top), batch (middle), and cluster (bottom).
    }
    \label{fig:umap}
\end{figure*}

\section{Limitations \& Future Work}
\label{app:lim_fut}
Having shown that scTree is equipped to discover hierarchical structures, there are still many interesting avenues to explore. Finding a stopping criterion is an exciting question, as the ground-truth number of clusters is usually unknown. Regarding the model architecture, finding a way to reduce the number of hyperparameters to tune would be beneficial. We believe our proposed configuration in Appendix~\ref{sec:hyperparameters} can serve as a good starting point for this. Similarly, the method is currently restricted to a binary tree, which could be generalized to better represent cell type hierarchies. Furthermore, as each leaf has a separate embedding, there is only the root embedding representing all samples, which hinders simple interpretations of the latent space(s). Lastly, NMI$_\text{batch}$ shows that scTree sometimes does not full correct for batch effects, and regularizing the learned representations more explicitly to prevent this might increase clustering performance even more.

\end{document}